\documentclass{article}

\usepackage{arxiv, placeins}

\usepackage[utf8]{inputenc}
\usepackage[T1]{fontenc}
\usepackage{hyperref}
\usepackage{url}
\usepackage{booktabs}
\usepackage{amsfonts}
\usepackage{nicefrac}
\usepackage{microtype}
\usepackage{lipsum}

\usepackage[
    backend=biber,
    style=ieee,
  ]{biblatex}
 
\usepackage{graphicx}  
\usepackage{subcaption}
\usepackage{booktabs}
\usepackage{mathtools}
\usepackage[flushleft]{threeparttable}
\usepackage{url}
\usepackage{authblk}

\graphicspath{ {./figs} }
    
\title{Sameness entices, but novelty enchants in fanfiction online}
\author[3]{Elise Jing}
\author[2]{Simon DeDeo}
\author[1,4]{Devin Robert Wright}
\author[1,5]{Yong-Yeol Ahn}
\affil[1]{Center for Complex Networks and Systems Research, Luddy School of Informatics, Computing, and Engineering, Indiana University Bloomington}
\affil[2]{Social and Decision Sciences, Dietrich College, Carnegie Mellon University}
\affil[3]{Sirius XM}
\affil[4]{Cognitive Science Program, Indiana University Bloomington}
\affil[5]{Indiana University Network Science Institute, Indiana University Bloomington}

\begin{document}

\maketitle
\begin{abstract}
Cultural evolution is driven by how we choose what to consume and share with others. A common belief is that the cultural artifacts that succeed are ones that balance novelty and conventionality. This balance theory suggests that people prefer works that are familiar, but not so familiar as to be boring; novel, but not so novel as to violate the expectations of their genre. We test this idea using a large dataset of fanfiction. We apply a multiple regression model and a generalized additive model to examine how the recognition a work receives varies with its novelty, estimated through a Latent Dirichlet Allocation topic model, in the context of existing works. 
We find the opposite pattern of what the balance theory predicts---overall success decline almost monotonically with novelty and exhibits a U-shaped, instead of an inverse U-shaped, curve.
This puzzle is resolved by teasing out two competing forces: sameness attracts the mass whereas novelty provides enjoyment. 
Taken together, even though the balance theory holds in terms of expressed enjoyment, the overall success can show the opposite pattern due to the dominant role of sameness to attract the audience. 
Under these two ``forces'', cultural evolution may have to work against inertia---the appetite for consuming the familiar---and may resemble a punctuated equilibrium, marked by occasional leaps.
    \end{abstract}


\section*{Introduction}
    \noindent 
    The culture industry may strive constantly for the latest hit~\cite{adorno}, but even with extensive research and strong financial incentives, it remains difficult to predict the success of a film, book, or television series~\cite{de2003hollywood}. One basic principle, however, is widely accepted: people enjoy familiarity~\cite{mull1957effect}. This preference is found in a wide variety of domains. Humans, for example, find faces that are close to the average, and therefore more familiar, more attractive. The effect extends to images of birds and automobiles~\cite{Halberstadt2003}, and even to arbitrary visual patterns: a classic psychological experiment by Zajonc~\cite{zajonc1968attitudinal} showed that mere exposure to a certain stimuli can increase people's preference for it, even when they are not aware of the exposure~\cite{kunst1980affective,bornstein1989exposure}. 
    
    These work lead naturally to the idea that successful cultural products leverage the familiar. Repetition has long been a central feature of music and poetry~\cite{huron2013psychological}, and the trend persists in contemporary popular culture, where new releases are often adaptations, remakes, and remixes of existing works~\cite{manovich2007comes}. Marvel and DC Comics have successful movies based on the familiar characters now decades old, including sequels, prequels, and reboots. The origin story of Spider-man provides an extreme case; this single narrative has been re-made, in slightly different forms, in three movies in the last eighteen years: \emph{Spider-Man} (2002), \emph{The Amazing Spider-Man} (2012) and \emph{Spider-Man: Into the Spider-Verse} (2018)~\cite{spider}, and all three movies achieved high recognition. In 2021, eight of the ten highest grossing movies were adaptions, sequels, remakes, or parts of a movie universe rather than \emph{de novo} creations~\cite{2021film}; in 2022, nine of ten were.~\cite{2022film}.
        
    Repetition, however, is not all. Extremely novel cultural products sometimes also enjoy huge successes. Popular music in the 20th Century is characterized not only by the stability of genres, but by high rates of turnover and the emergence of new ones~\cite{mauch2015evolution}. Widely-imitated genres such as hip-hop and electronic dance music began in small underground communities and grew, despite their initial unfamiliarity, to enjoy global dominance. In the film industry, successful but novel products include not just new characters and situations, but also new forms of cinematography. In the history of literature, experimental works such as \emph{The Waste Land} by T.S.\ Eliot, \emph{The Waves} by Virginia Woolf, and \emph{Ulysses} by James Joyce were unexpected events that reorganized the canon and defined an era. This poses a challenge for those who study cultural evolution, which must account at once for both the known psychological preferences for what is familiar, and the success of these famous counterexamples.
    
    \textbf{`Balance' Theory of Liking}. A widely-adopted hypothesis reconciles the apparent conflict between the preference for novelty and familiarity by suggesting that successful creative works are a combination of, or balance between, convention and innovation. According to this theory, popular works are different from previous works and their peers, but not \emph{too} different. In psychology, this idea is captured in the Wundt-Berlyne curve, an inverted-U curve with an optimal amount of novelty for hedonic values at an intermediate position~\cite{berlyne1970novelty}; similar accounts are found in the business and marketing literature where it is known as Mandler's Hypothesis~\cite{meyers1989schema}, and in organizational theories as the optimal distinctiveness hypothesis~\cite{zuckerman2016optimal}.
    
    A few experiments have supported this hypothesis in the case of words~\cite{sluckin1980liking}, music~\cite{hargreaves1984effects,askin2017makes}, films~\cite{sreenivasan2013quantitative}, advertising~\cite{mohanty2016visual}, and scientific publications~\cite{uzzi2013atypical}. These results suggest that, for example, songs with a non-zero, but non-extreme, level of differentiation are more likely to be on the top of the Billboard's Hot 100 charts, that movies balancing familiarity and novelty have higher revenues, or that visual metaphors in ads work best when they have mild incongruity. In scientific publications, the highest-cited papers are argued to be grounded on mostly conventional, but partly novel combinations of previous works~\cite{uzzi2013atypical}. 
    
    However, much of these research deals with cultural products whose reception are strongly influenced by factors beyond intrinsic ones such as content, style, subject, genre, or length. External factors that complicate these analyses include price, advertisement, and media coverage. A song's popularity may be only partially determined by its quality~\cite{salganik2006experimental}, and deliberately putting unpopular songs on the top chart can in fact popularize them~\cite{salganik2008leading}. Indeed, belief in an inverted-U can be self-fulfilling: a scientific funding agency or a film company may choose to give critical funding to work of incremental novelty in the expectation that it will succeed. In addition, the modern culture industry is concerned with multiple metrics for success; a movie that begins with a high box office may not earn a good rating among its audience, and a highly celebrated work may not have an equally successful sequel. The complex interactions between consumers, cultural products, and the market is difficult to disentangle.

    \textbf{Fanfiction and Fandoms}. In order to test the balance theory while controlling for external effects, we study an unusually rich dataset of fanfiction work that allows us to isolate the effect of novelty. Fanfiction draws on narratives and characters from a canonical work to create new stories and alternate timelines~\cite{wiki:transf_work}. Anhttps://www.overleaf.com/project/5cb3922f51f9726ae110835a enthusiast of the Harry Potter series may write a new adventure for Hermione and Harry; a fan of the television show Buffy the Vampire Slayer may write a story in which Willow's girlfriend has a different fate. 
        
    Despite being based around a canonical work, fanfiction is one of the most innovative practices in contemporary culture, turning people who would, ordinarily, only be consumers, into creators. It is playful, transformative, and transgressive~\cite{ff1,ff3}: fans of Sherlock Holmes, for example, have written stories in which Holmes and Dr.~Watson fall in love, and Watson, by magical means, gestates the couple's baby. Examples like these abound, and fanfiction communities (``fandoms'') are far from conservative, often subverting, as well as extending, a canonical work. This production goes along with an interchangeability of the creator and consumer roles: an author also reads and comments on other fanfiction~\cite{eiji2010world}. These interactions happen on time scales of hours, even minutes, and the rapidity of the feedback process allows for rapid change and selection, making them ideal laboratories for the study of cultural evolution.
        
    The nature of fanfiction also allows us to mitigate a number of common confounds in the study of creativity and what people enjoy. First, fanfiction is usually shared within a fandom's community with no advertisement or promotions \cite{wiki:fandom,ff2}; a work's success is therefore largely uninfluenced by top-down interventions, such as advertising campaigns, that can distort its reception. Second, most fanfiction works are freely available on the Internet, so that the price is not a confounding factor. Third, works in the same fandom are created within the same context, and have similar subjects, characters, and settings to each other. As in genres of literary production, variations between works occur in a recognizable space, allowing us to talk about their novelty while controlling for other factors. Because fanfiction is available in plain text and in remarkably large volumes, systematic, statistical approaches are possible. Finally, our data is rare in that it records both the viewership (number of hits) and the enjoyment (number of Kudos; see Methods), allowing a deep look into the complicated process of audience feedback.
    
    We draw our data from the online fanfiction archive Archive of Our Own (AO3; see Methods). In total, we collected 671,908 pieces of fanfiction in 23 fandoms from 124,305 authors and study the relationship between a fanfiction's novelty and its success. We use a topic model (Latent Dirichlet Allocation; LDA) to characterize the novelty of fanfiction (a term-level novelty measurement showed similar results; see Supplementary Information). A work is evaluated with respect to the existing work in the same fandom; we consider it to be more novel if it is more different from the other fictions published during the previous time period in the feature space (see Methods). The success of a fanfiction is measured by four variables: hits (number of views), ``Kudos'' (similar to a `like' on Twitter or Facebook), comments, and bookmarks. Further metadata allows us to control for other effects. 
    
    Our results show an almost monotonically decreasing relationship between a fiction's novelty and its success. Furthermore, our regression models suggest that the relationship between success and novelty is U-shaped, rather than inverse-U-shaped. Readers seem to prefer fanfiction that give them a sense of familiarity, and we find no limit to their appetite for ``more of the same''; at the same time, high-novelty seems to have some chance to achieve success.
    As we will explain below, this result that seemingly contradicts the balance theory can be, however, resolved by teasing our two factors in play: although novelty-enjoyment relationship still follows the balance theory, the familiarity's attractiveness dominates the overall success.  
    Taken together, the relationship between novelty and enjoyment follows the traditional ``inverted  U-shape'' account, but the relationship between novelty and attraction reverses it.

    \section*{Data \& Methods}

    \textbf{Archive of Our Own}
    AO3 is an online archive which allows users to upload their fanfiction, and categorizes them based on fandoms. Established in 2009, it has become one of the largest fan communities, with more than 1.6 million users and 4 million works by August 2018~\cite{AO3stats}. A Python script was used to download all fanfiction and their metadata from AO3 (\url{http://archiveofourown.org/}) in our selected fandoms (see below) in March 2016. 

    AO3 classifies the fandoms based on the formats of the canons, such as movies, TV shows, books, anime \& manga, and musicals. We first identify the top five fandoms in each of these categories based on the number of works they contain. We exclude the fandoms that are subsets of other large fandoms. For example, we keep \emph{Marvel} but exclude \emph{The Avengers}, because \emph{The Avengers} is a part of the Marvel Universe. Fandoms that do not have a unified subject, such as K-pop (which contains fanfiction about over 300 different Korean pop bands), were also removed. In addition, we removed cross-over fanfiction which feature characters or elements from more than one fandoms. We keep only the fanfiction written in English. Finally, to control for the effects of very short works, we only analyze the subset of fiction with more than 500 words. 
        
    Fig.~S1(a) shows the number of works in each of the 23 fandoms; \emph{Marvel}, \emph{Supernatural}, and \emph{Sherlock Holmes} are the largest. Fig.~S1(b) shows the volume of works produced over time. AO3 was established in December 2009, and experienced rapid growth beginning in 2012. Because the works timestamped earlier than the start date might be migrated from other platforms, and may not correctly reflect the status of the archive, we only run our analysis using those published in January 2010 or later.

    The success of a fanfiction is measured by its hits, Kudos, comments, and bookmarks. While the number of hits is the most direct metric for popularity, the number of Kudos is a clearer signal that readers enjoy the fiction. A reader can also comment on a fiction or bookmark it to read later. The comments and bookmarks therefore signal the recognition or engagement from readers, although they depend on multiple motivations, and are less directly associated with popularity.
     
    Fig.~S2 shows the distribution of the hits, Kudos, comments, and bookmarks on a logarithmic scale. Like many other measurements of popularity, they exhibit fat tails, with a small number of works receiving the majority of attention and most receiving little or none. Also note that there are outliers that achieved extreme recognition in terms of hits and Kudos. In the analysis, we log-transform the response values values, a common practice for similar data such as citations \cite{thelwall2014regression}. This practice has been argued to reduce the potential bias when performing regression and other statistical analysis \cite{thelwall2014regression}. The metadata that we collected for each fanfiction is summarized in Table \ref{tab:metadata}.

    \textbf{Quantifying novelty}. Although there are many ways to measure novelty, here we operationalize it in an intuitive, data-driven way. As previous studies noted \cite{askin2017makes,de2015game}, it is reasonable to assess a work's novelty in the context of other previously published works. Intuitively, a work is less novel if it is similar to many others published beforehand. Here we use the centroid, in feature space, of all past works in a fandom during a certain period as the guide for measuring novelty. A work is more novel the further it is from the center. (See Supplementary Information for more details about our operationalization of novelty.)

    In line with many existing researches such as \cite{klingenstein2014civilizing,barron2018individuals,horvat2018role}, we extract features from the works using the Latent Dirichlet Allocation (LDA)~\cite{blei2003latent}, characterizing documents in terms of topics, or co-occuring word patterns. We construct document-term matrices with each fanfiction work as a row, and each term in the vocabulary as a column. For each work, we fit a LDA model on all fanfiction published in the same fandom within the past 6 months from when it was published. We therefore construct a feature space consisting of the topic distributions of all works from this time period. We then compute the centroid of the feature space as the average of all feature vectors. The novelty score $s_{i}$ is defined as the Jensen-Shannon Distance (JSD; \cite{klingenstein2014civilizing}) between the fanfiction's topic distribution and the center of the feature space:
        
    \begin{equation}
    \begin{centering}
    s_{i} = \frac{1}{2}D(\boldsymbol{f_i^{(c)}}\rvert\rvert\boldsymbol{v}) +\frac{1}{2} D(\boldsymbol{f_i}\rvert\rvert\boldsymbol{v})
    \end{centering}
    \end{equation}

    where $\boldsymbol{f_i}$ is the vector representation of a fanfiction, and $\boldsymbol{f_i^{(c)}}$ is the centroid (see above). $\boldsymbol{v} = \frac{1}{2}(\boldsymbol{f_i} + \boldsymbol{f_i^{(c)}})$, and D is the Kullback--Leibler divergence between the two vectors.
        
    The Python library \texttt{gensim} is used to fit LDA topic models on our data. The parameters are: number of topics $=$ 100\footnote{Our results do not significantly change even if we vary the number of topics to 20, 40, 60, 80, and 120 (results not shown).}, $\alpha = 0.01$, and iterations $=$ 50. The texts are preprocessed by removing the top 500 most frequent words and the words that appear only once; a common practice in fitting LDA models~\cite{gensim_tutorial}. Additionally, we remove the punctuations, numbers, and convert all text to lower case. The data and code that we used is made available\footnote{\url{https://github.com/yzjing/ao3}}.
    
\section*{Results}
    
    We first sense-check our operationalization of novelty by qualitative checks and close-reading. Examining individual fanfiction with different levels of novelty, we found that the least-novel fictions (in the 95\% percentile) and intermediate-novel fictions (around the 50\% percentile) tended to feature common character pairings and story settings. Meanwhile, top-novelty fanfictions (in the top 5\% percentile) often feature unusual writing styles, story elements, or character pairings. 
    One example---the most extreme, in stylistic innovation---is the work \emph{I am Groot} from the \emph{Marvel} fandom. ``Told from the perspective of Groot,'' it has a novelty score among the top 0.1\%. The fiction consists of 437 repetitions of a single sentence ``I am Groot.'' Similarly, a high-novelty fiction in the \emph{Sherlock} fandom, titled \emph{The Real Meaning of Idioms}, is written entirely as exchanges of text messages. Our novelty measures are unsupervised, but appear to be successful in capturing a range of intuitively innovative projects.

\begin{figure}
    \centering
    \includegraphics[width=\linewidth]{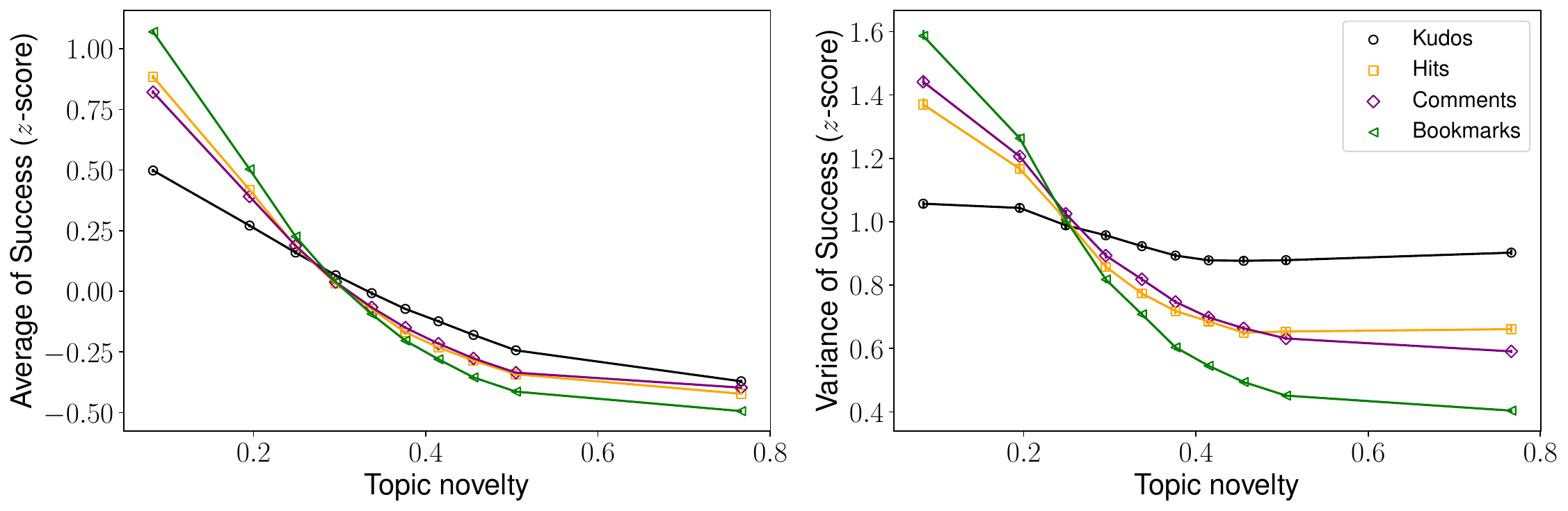}
    \caption{The interaction between novelty score and success. The engagement metrics (hits, Kudos, comments, and bookmarks) and its variance tend to decline although the variance of Kudos does not decline (or even slightly increases) high-novelty works. The horizontal axes are the novelty scores divided into percentiles. The left column shows the corresponding average of the $z$-score of Kudos, hits, comments, and bookmarks, and the right column shows the variance. 95\% confidence intervals obtained from bootstrap resampling are shown.}
    \label{fig:lda_kudos}
    \end{figure}

    \textbf{Correlation between novelty and success}. Figure~\ref{fig:lda_kudos} shows the interaction between the novelty score and hits, Kudos, comments, and bookmarks (see Methods). We show the average and variance of these metrics, aggregated across fandoms, as novelty increases. In addition, Figure \ref{fig:kudos_term_only} shows the same interaction using a term-level novelty measurement (see Supplementary Information), where we observed a similar pattern. Because of the wide dynamic range and fat-tailed distributions (see Supplementary Information), we compute the log values of the success metrics; since the fandoms differ in size and amount of fan activity, we present the $z$-scores of these values. As the topic novelty score increase, the average $z$-scores of responses in the form of hits, Kudos, comments, and bookmarks decrease across the board, displaying a negative correlation. The more novel a work becomes, in short, the less likely readers read, engage, or express enjoyment of it.

\begin{figure}[!htb]
    \centering
    \includegraphics[width=\linewidth]{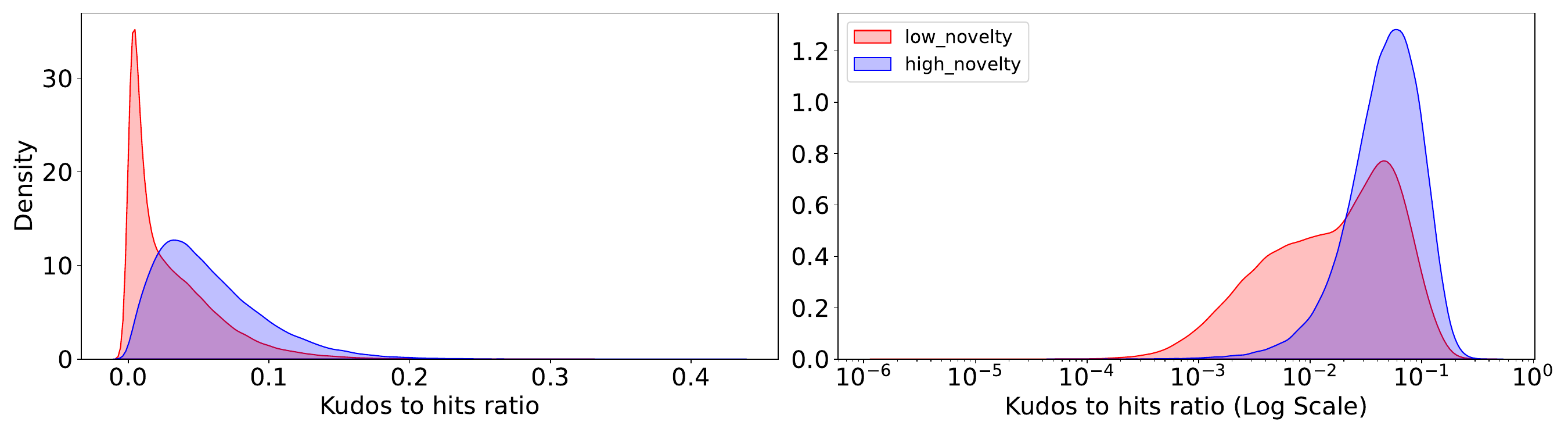}
    \caption{Kernel density estimation (KDE) of the distribution of Kudos-to-hits ratio for fanfiction in the top 25\% and bottom 25\% novelty range (Left: linear scale; Right: log-scale). High-novelty fanfiction are observed to receive more kudos with respect to the number of hits they receive, although having fewer hits overall. This pattern can also be observed for every fandom individually (see Supplementary Information).}
    \label{fig:agg_kde}
    \end{figure}

    Because engagement (e.g., comments and Kudos) is conditional upon the reading of the piece, we also examine the ratio between Kudos and hits, which captures the likelihood of expressing enjoyment, conditioned upon the reading of the piece. 
    For instance, it is possible that high-novelty works receive more enthusiastic response but only from a small niche audience. 
    To test this, we compare the ratio between Kudos and hits for fanfiction in the low novelty range (bottom 25\%) and high novelty (top 25\%). 
    A higher ratio indicates that a larger percentage of people who read the work have left Kudos on it. Figure~\ref{fig:agg_kde} shows the kernel density estimation of the distribution of Kudos-to-hits ratio for the low and high novelty works. We found that the high novelty works have a significantly higher Kudos-to-hits ratio.
    This pattern holds for most fandoms when they are analyzed separately (See Fig.\ref{fig:kde_indiv}), but is not observed in our term-level novelty measurement. 
    In other words, although high-novelty works tend to be read by fewer people, those who read tend to like it more. 
    This observation motivates us to consider the Kudos-to-hits ratio as another key metric that captures enjoyment. 
    In order to better isolate the relationship between the novelty and success, we conduct a regression analysis in the next section.

    \subsection*{Regression analysis}

    \textbf{Response variables}. We consider the same five response variables (log of Kudos, hits, comments, bookmarks, and Kudos-to-hits ratio) in our regression models.

    \textbf{Independent variables}. The topic novelty score is the predictor variable in all models. To account for possible non-linear relationships such as the inverted U-shape curve, we also use the square value of the score as an additional predictor variable. We mean-center the topic novelty score before computing the square value. 

    \textbf{Control variables}. We consider the following control variables. (1) fandom fixed effects; (2) number of chapters: multi-chapter works have additional chances for exposure, and higher metrics may stimulate an author to write more chapters; (3) time of publication: AO3's user base has been increasing and one may expect either newer works to receive higher metrics than older ones because they are visible to a larger audience, or that older works may perform better because they had more time to accumulate readers. We use the number of days since a work was completed (for finished works) or was last edited (for incomplete works); (4) total number of works by the author in question: authors may accumulate fame which may bias readership and success; (5) character-pairing: the relationship between characters is one of the main reasons that many fans read fanfiction, and some relationship pairings have larger fan bases than others. To account for this effect, we control for a binary variable, ``frequent relationship'', to indicate whether a work features a relationship that is among the top five most frequent relationships in its fandom. Finally, (6) the ``archive warnings'' indicate that the fanfictions contain sensitive elements such as graphic violence or major character death, and may influence the readers' choice to read them; the age ratings restrict some fictions to adults only; the types of character relationships may also influence the readers' choices. Categorical variables are created to capture their effects. 

    \begin{table}
    \centering
    \begin{tabular}[width=0.5\linewidth]{p{0.6cm}p{3cm}p{2.5cm}p{1.5cm}}
    \toprule
    Model & Response variables & Independent variables & Control variables \\ 
    \hline
    1-4 & Logarithm of Kudos, hits, comments, bookmarks, and Kudos-to-hits ratio respectively & Topic novelty & All control variables \\
    5-8 & Logarithm of Kudos, hits, comments, bookmarks, and Kudos-to-hits ratio respectively & Topic novelty, square value of topic novelty & All control variables \\
    9-12 & Non-zero subsets of Kudos, hits, comments, bookmarks, and Kudos-to-hits ratio & Topic novelty & All control variables \\
        
    \bottomrule
    \end{tabular}
    \caption{The response, independent, and control variables used in each group of the regression models.}
    \label{tab:reg}
    \end{table}
        
\begin{figure}
    \vspace{-1.1cm}
    \centering
    \includegraphics[width=\linewidth]{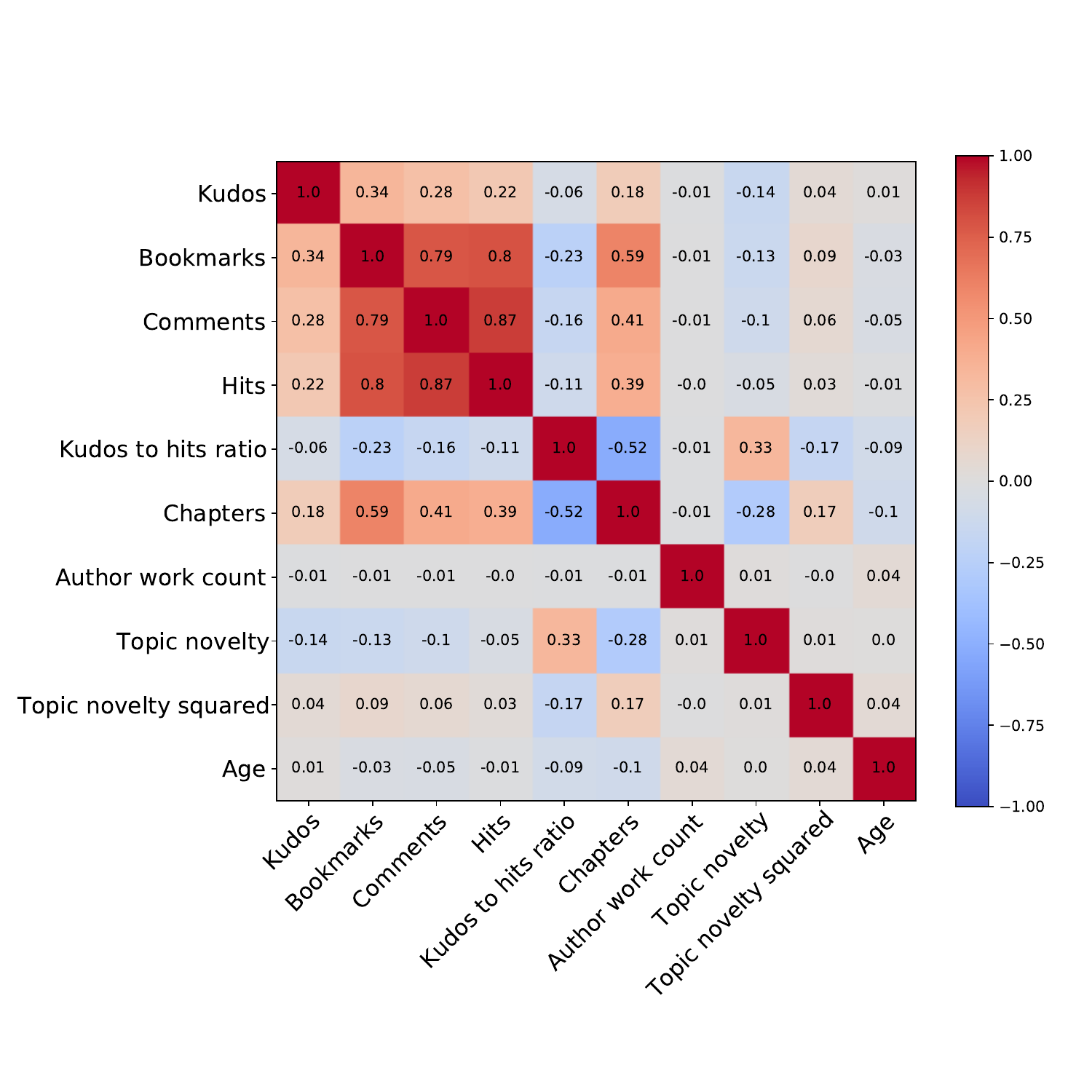}
    \vspace{-1.1cm}
    \caption{Correlations between the numerical predictor, response, and control variables. Strong positive correlation is found between the response variables except for the Kudos-to-hits ratio. Topic novelty is weakly positively correlated with Kudos-to-hits ratio, but negatively correlated with the other response variables. Additionally, the number of chapters is moderately positively correlated with success.}
    \label{fig:corr}
    \end{figure}

    The correlations between the numerical variables are shown in Fig.~\ref{fig:corr}. We noticed weak negative correlations between the novelty score and hits, Kudos, comments, and bookmarks; however, the novelty score is positively correlated with Kudos-to-hits ratio.  Moderate to strong positive correlations exist between the number of chapters and bookmarks/comments. However, no strong pairwise correlation is found within the predictor and the numerical control variables. We also examined the Variation Inflation Factor (VIF) and removed one variable that causes strong collinearity (the age of the work), although keeping it does not qualitatively influence the results. The variables that each model contains are summarized in Table~\ref{tab:reg}. 
        
    Because there are many zero values in the response variables (many works receive no comments at all, for example), we use a two-part model~\cite{jones2000health,humphreys2013dealing} (we do not use zero-inflated Poisson or negative binomial regression models because our outcomes are the average of log-counts data). A logistic regression is first performed on the predictor and control variables to predict the probability of each sample having a non-zero outcome. This probability is then used as an additional predictor variable in a pooled OLS regression on the samples \emph{with non-zero outcome}\footnote{Simply discarding or keeping all zero responses do not qualitatively change our model outcomes (results not shown).}. We use the Python library \texttt{statsmodels} \cite{seabold2010statsmodels} to perform the analysis.
        
\begin{figure*}[h]
    \centering
    \includegraphics[width=\textwidth]{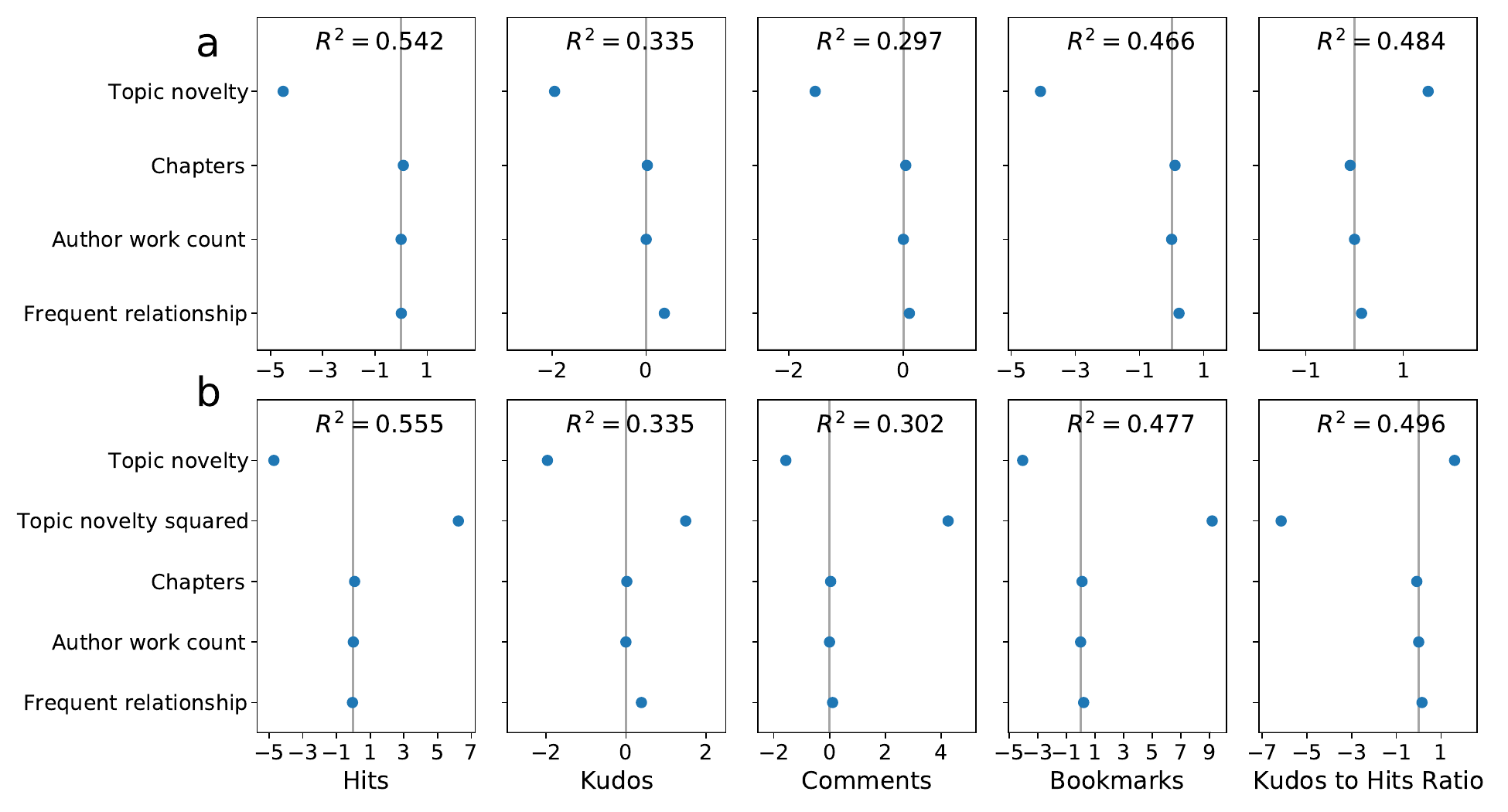}
    \caption{OLS coefficients for the independent variables and selected control variables for the multiple regression models. $N = 520,730$. Figure (a) does not include the square value of novelty score, while Figure (b) does. 95\% confidence intervals are shown. The coefficients of the categorical variables are omitted (see Supplementary Information for the full coefficients). Among the control variables, the number of chapters and the author's fame does not have an effect on the outcome, while featuring a frequent relationship has a positive effect. In both Figure (a) and (b), novelty has negative effects for hits, Kudos, comments, and bookmarks; however, Figure (b) shows that the squared value of novelty has a large positive effect on these outcomes. The effect on Kudos-to-hits ratio exhibits the opposite pattern.}
    \label{fig:ols_partial}
    \end{figure*}

    Selected OLS coefficient estimates of the models are shown in Figure~\ref{fig:ols_partial}. We consider the control variables first. A work featuring frequent relationships is slightly more likely to have better reception. Contrary to our initial assumptions, the number of chapters and the author's fame do not have significant influence on success. While the coefficients of the categorical control variables are not shown here, we found that work published in newer and more popular fandoms such as \emph{Star Wars} and \emph{Marvel}\footnote{These fandoms may have long histories, but recent installments such as Star Wars' new trilogy and the Marvel movies are associated with the influx of new fans.} tend to be more successful. Elements such as character death and violence are associated with poorer recognition. For full results with all coefficients, see Fig.~S3.
        
    \begin{figure*}
    \centering
    \includegraphics[width=\textwidth]{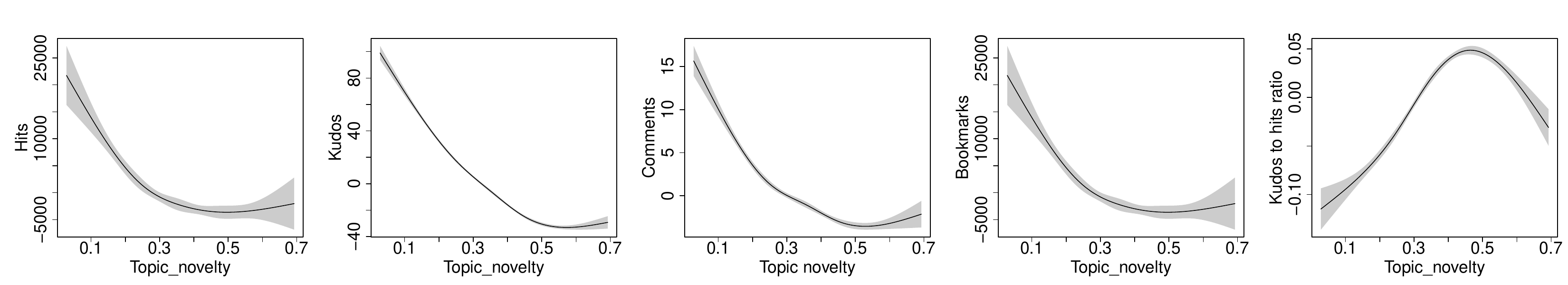}
    \caption{Models 9-12: Partial dependence plots of the Generalized Additive Models. The x-axes show the novelty scores. The y-axes show the partial influence of the novelty score on the response variables while holding other independent variables constant. 95\% confidence intervals are shown.}
    \label{fig:gam}
    \end{figure*}
    
    We then examine models 1--4, which do not include the square value of the novelty score (Figure~\ref{fig:ols_partial}(a)). Upon controlling our control variables, we find that novelty has a negative effect on Kudos, hits, comments, and bookmarks, but a small positive effect on Kudos-to-hits ratio, supporting our previous observation that higher novelty is linked to lower success but more enjoyment. For example, increasing novelty by 0.1 (cf. Figure \ref{fig:lda_kudos}) is associated with the decrease in Kudos by approximately 11.8\% . 
        
    When we add the squared values of novelty scores in models 5-8 (Figure \ref{fig:ols_partial}(b)), the coefficients of the novelty score are similar to that in models 1-4, suggesting the robustness of our models. At the same time, we find that the effect sizes of the squared novelty score are all positive and large for Kudos, hits, comments, and bookmarks. Both fanfiction with low and high novelty are therefore associated with better reception, suggesting not the \emph{inverted} U-shaped curves, but \emph{U-shaped} curves. Meanwhile, the coefficient for Kudos-to-hits ratio suggests an inverted U-shaped curve. We observe similar results with our term-level novelty measurement (see Figure \ref{fig:ols_term_only}).
    
    \textbf{GAM.} Our regression results suggest a nonlinear relationship between novelty and success. To examine this further, we turn to generalized additive models (GAM)~\cite{hastie2017generalized}, which allows us to study non-linear relationships in complex data more directly (e.g.~\cite{horvat2018role}). We use the non-zero subset of the response variables now without log transformation, and use the same predictor and control variables as in the linear regression models (see Table~\ref{tab:reg}). The models are fit using the \texttt{mgcv} library in \texttt{R}, with the following parameters: $ k = 5 $ and $sp = 0.1$. 
    
    The partial dependence plots of the models are shown in Figure~\ref{fig:gam}. Results show an overall decreasing relationship between novelty and the response values, but also suggests a higher uncertainty with the large confidence intervals, with upticks for very high novelty. These plots present a U-shaped curve, confirming our observation from the previous section. At the same time, the results for Kudos-to-hits ratio suggests an inverted U-shaped curve, where moderate novelty is associated with the highest outcome. 
    In other words, conditioned upon the reading, a fanfiction is more likely to receive kudos if it has a moderate amount of novelty, which concurs with the balance theory. Although we find similar patterns for Kudos, hits, comments, and bookmarks with our term-level novelty measurement (see Figure \ref{fig:gam_term_only}), the inverted U-shaped curve for Kudos-to-hits ratio is only observed with topic-level novelty.

    Together, our results capture distinct dynamics across the two stages of recognition. 
    The first stage is about readers' choice to click \& read. 
    Our results suggest that the \emph{familiarity drives engagement}---the more familiar it sounds, the more likely people would read it. 
    While this effect dominates, extreme novelty may also drive engagement as a U-shaped curve is suggested by the models. 
    This selection dynamics largely drive other engagement metrics as well. 
    The more people read it, the more Kudos, bookmarks, and comments it will likely receive. 
    However, there may be another important process in play: the \emph{novelty drives enjoyment}, although too much novelty may reduce the enjoyment (inverted U-shape suggested by the regression models). 
    The final outcome of audience appreciation is the result of these two processes that are working in the opposite direction---sameness entices, but novelty enchants.

\section*{Discussion}
    Traditional theories suggest that people like things with a balance of familiarity and surprise. Our findings from the fanfiction community further develops these theories by disentangling two metrics for success: attraction and enjoyment. Attraction is found to follow a U-shaped curve (with a decreasing trend) while enjoyment follows an inverted U-shaped curve (with an increasing trend).

    The boundaries that fandoms impose on themselves provide a natural control for the variation in subjects, characters and settings, allowing us to better isolate the influence of novelty, and avoiding the confounding factors that may have contributed to the single inverted U-shape curve found by previous studies.
    
    Our study has many limitations. First, our study design does not allow us to draw causal relationships between our variables. Our results are based on observational data analysis and thus correlational.

    Our study is also focused on a specific genre of cultural products that are produced and consumed as a subculture. 
    Although online fandoms are well-isolated and better-controlled system, both the authors and the consumers of the works are drawn from a particular sub-population skewed towards young females, and largely from English-speaking countries. Fanfiction is usually the domain of amateur writers whose training, socialization, and incentives may differ from the ``professional'' producers of cultural products in other domains. 
    Fanfiction is also a special genre.
    While all cultural practices have a chain of inheritance, fanfictions are more explicitly anchored in original texts than most as they are based on canons, and usually include established characters, personalities, world settings. The readers of fanfiction may thus have a bias towards the familiar.
    Therefore, our results may not generalize well to other types of cultural products. 
        
    Our characterization of novelty captures word usage and topic variations (see Methods); while we find similar patterns for attraction with our term-level and topic-level novelty measurement, the optimal differentiation phenomena is much stronger with topic-level novelty measurement. Both methods, however, neglects the semantic information contained in word orderings. In literary theories, the arrangement of events plays an essential role in stories. Our methods capture the ``material'' of stories but are unable to evaluate the way it is arranged. Moreover, the novelty of a piece of writing may appear in its style as well as in content. Some stylometric features, such as the usage of certain words, are captured by our methods, but they could not be decomposed from contents. Other features such as sentence length and punctuation usage are neglected by our methods. This treatment of stylometric features may bias our evaluation of novelty.
        
    Our results may also diverge from previous researches because of the methods we used. In the early experiments by Berlyne~\cite{berlyne1970novelty}, they controlled the novelty of geometric shapes by having the subjects exposed to them, and then evaluate the success by asking the subjects to report their ``interestingness''. Similarly, Zajonc's experiments exposed subjects to groups of words~\cite{zajonc1968attitudinal}. Such different data and setup can be one reason for the different outcomes. However, we also note that our results diverge from some recent researches that measure novelty and success similar to our study~\cite{horvat2018role,askin2017makes}, suggesting that our findings are not merely caused by the different experimental setups.
    
    Even with aforementioned limitations, our study may have important contributions and implications. 
    First, our analysis teases out two competing forces---attraction vs. enjoyment---and the contrasting roles played by familiarity and novelty, thereby providing a useful framework to investigate the enjoyment, success, and evolution in culture. 
    Second, as mentioned earlier, fanfiction is a well-controlled ecosystem that is free from many confounding factors that make studying mainstream cultural products tricky. 
    Third, our analytic strategies may be broadly applicable to other types of text narrative data. 
    Future work may identify causal inference study designs to establish the link between novelty and success of cultural products, or investigate other ways to quantitatively estimate novelty in cultural products.

    \section*{Acknowledgments}
    We thank the AO3 staff for helping us with the data collection, and thank Jaehyuk Park, \'Agnes Horv\'at, Sandra Ku\"ebler, Xiaozhong Liu, Minje Kim, and the anonymous reviewers for their comments.

\printbibliography

\appendix
    
    \section*{Appendix}
    
    \subsection*{Dataset}
\begin{figure*}
    \centering
    \hspace{1cm}            
    \includegraphics[width=\textwidth]{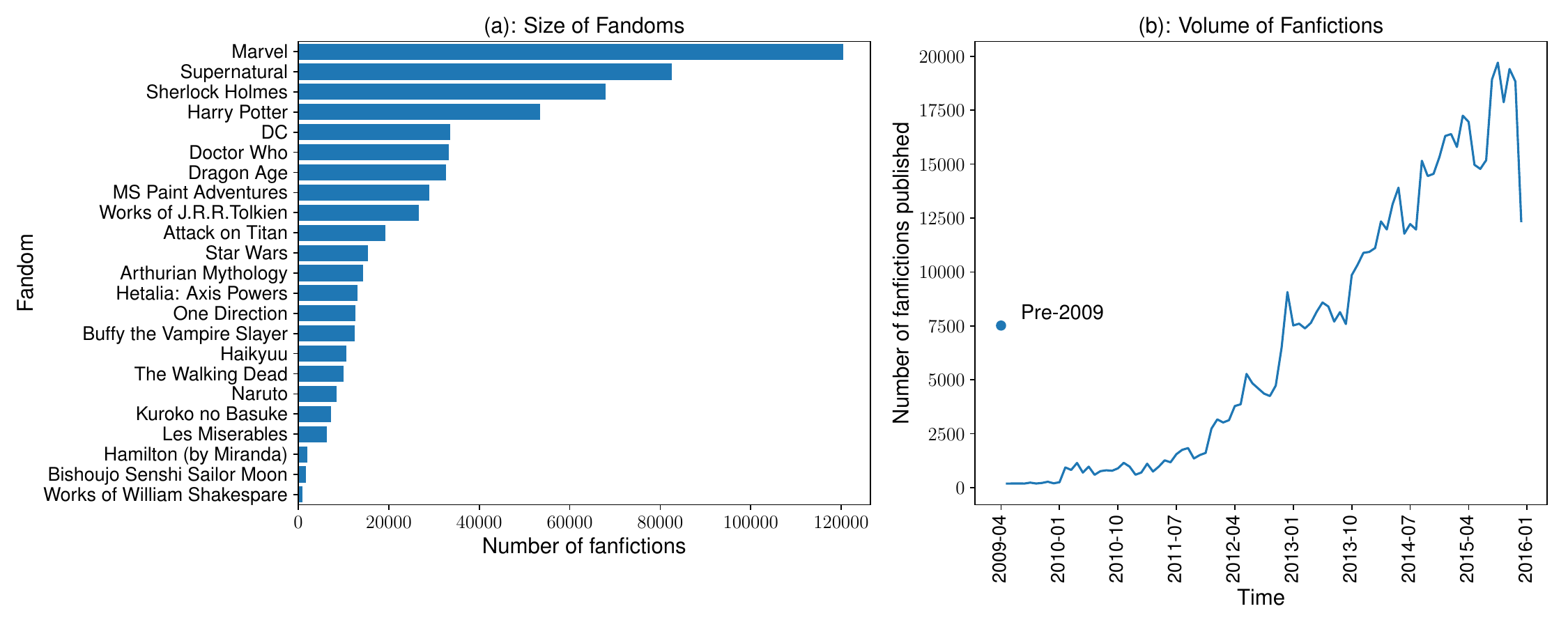}
    \caption{The size of fandoms and the amount of fanfiction published in AO3 each month, from 2009 to 2016.}
    \label{fig:fandom_size}    
    \end{figure*}

\begin{figure*}
    \centering
    \includegraphics[width=\textwidth]{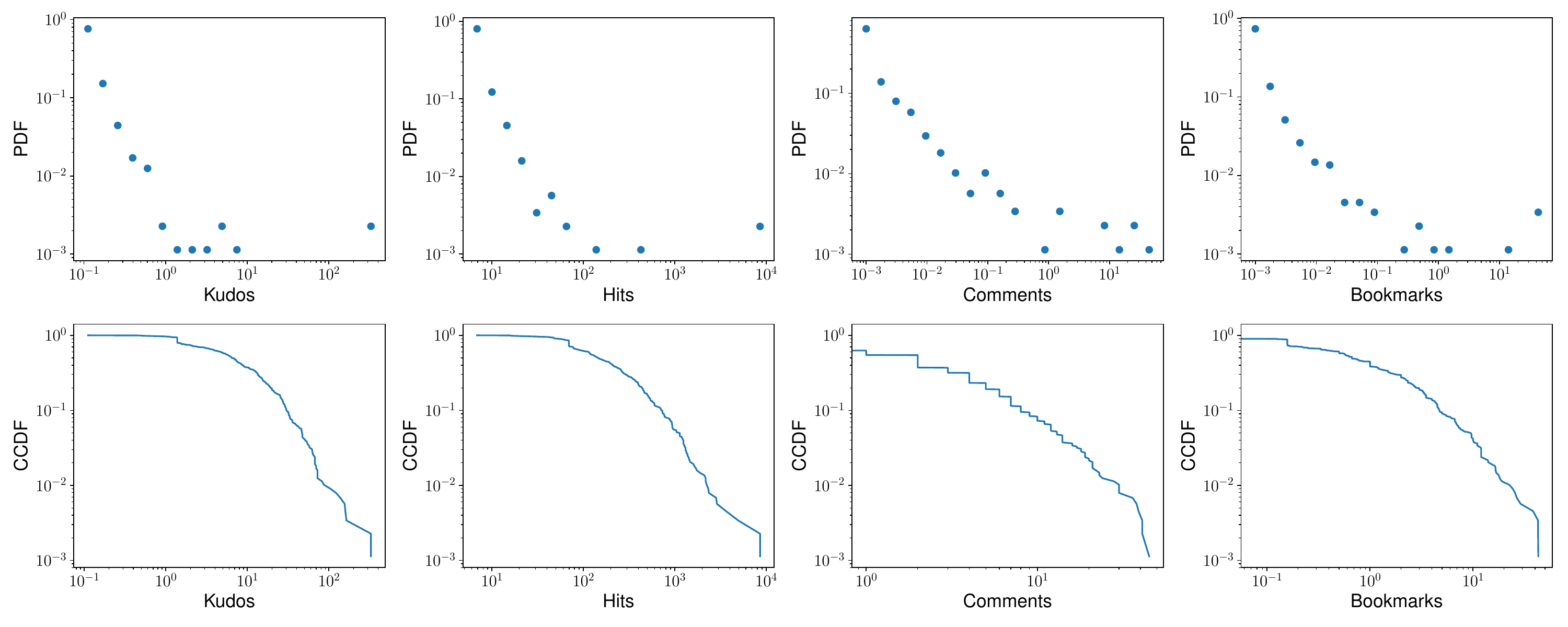}
    \caption{Log-binned probability density function and complementary cumulative distribution of kudos, hits, bookmarks and comments. Fat-tailed distributions are observed, where a small portion of fictions receive many Kudos, hits, etc., and most receive few.}
    \label{fig:kudos_dist}
    \end{figure*}

    Figure~\ref{fig:fandom_size} shows the distribution of fanfiction works across time and fandoms. The Archive experienced a steady growth since its founding in 2009, with the highest amount of works published in 2016. The largest fandoms include \emph{Marvel}, \emph{Supernatural}, and \emph{Sherlock Holmes}. Figure~\ref{fig:kudos_dist} shows the distribution of Kudos, hits, comments, and bookmarks in our dataset. Table~\ref{tab:metadata} shows the metadata fields available for each fanfiction piece.
    
    \subsection*{Data aggregation}
    While a fanfiction piece can have multiple chapters, the metrics for reception are for the whole work. To reconcile this conflict, we aggregate all chapters of the same work into a long document.

    \subsection*{Measuring novelty}
    To make sure that our results are not biased by the way we measure the novelty of text data, we experiment with different methods to define and evaluate novelty. 
    
    Our paper describes a topic modeling framework to measure novelty. Alternatively, we provide an approach to quantify novelty based on word frequencies. The term frequency--inverse document frequency (TF--IDF) is one the most fundamental and widely used ways to build a representation for documents. TF--IDF is a way of weighting co-occurence values of words often used in information retrieval~\cite{salton1988term}, where each document is represented as a vector, and its entries are the TF--IDF scores of the unique terms in the document. We first pre-process the texts by removing the terms that appear only once. The TF-IDF scores are then computed, for each fanfiction, using all fanfictions published in the same fandom within the past 6 months from when it was published. The Python library \texttt{scikit-learn} was used to create the vectors. We then compute the centroid of the feature space as the average of all feature vectors. The \emph{term novelty} score $s_{i}^{(\mathrm{term})}$ of a fanfiction is defined as:

    \begin{equation}
    s_{i}^{(\mathrm{term})} = 1-\frac{\boldsymbol{f_i} \cdot{\boldsymbol{f_i^{(c)}}}} {\lvert \boldsymbol{f_i} \rvert \lvert \boldsymbol{f_i^{(c)}} \rvert}
    \end{equation}
        
    where $\boldsymbol{f_i}$ is the vector representation of a fanfiction, and $\boldsymbol{f_i^{(c)}}$ is the centroid of the vector space defined with respect of this fanfiction.
    
\begin{figure*}
    \centering
    \includegraphics[width=\textwidth]{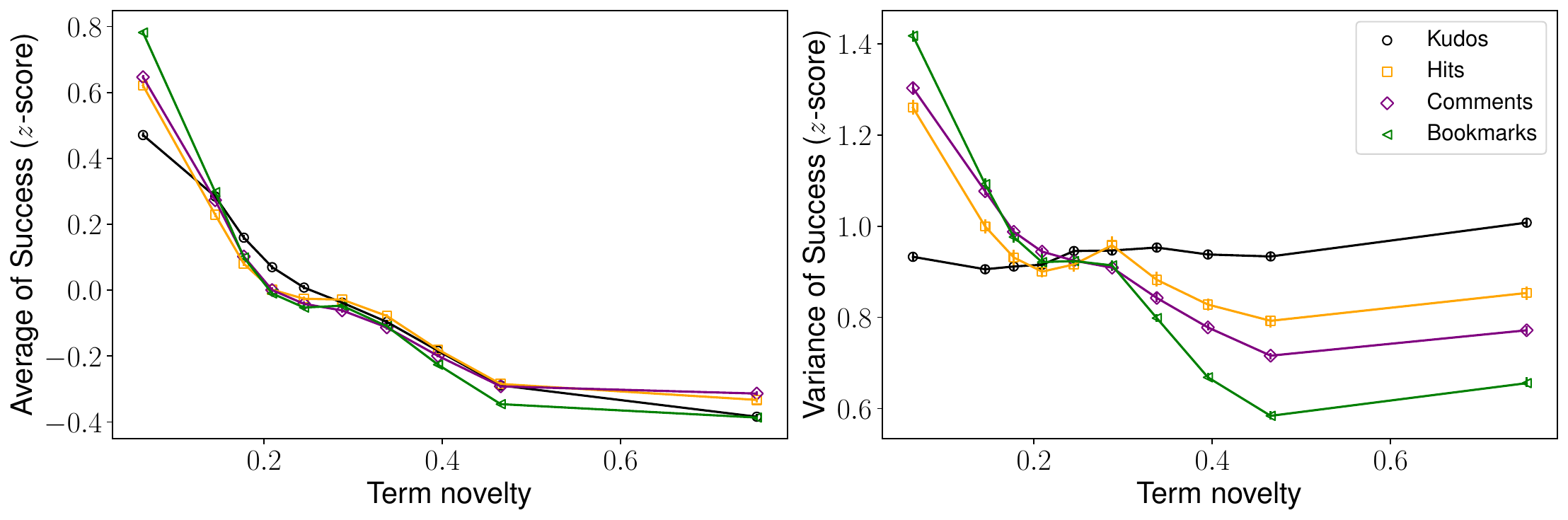}
    \caption{The interaction between term-level novelty score and success. Average returns decline, while variance declines(for comments and bookmarks) or slightly rises (for Kudos and hits), as novelty increases. Here we plot the average and variance of Kudos, hits, comments, and bookmarks. The horizontal axes are the term-level novelty scores divided into percentiles. The left column shows the corresponding average of the $z$-score of Kudos, hits, comments, and bookmarks, and the right column shows the variance. 95\% confidence intervals obtained from bootstrap resampling are shown.}
    \label{fig:kudos_term_only}
    \end{figure*}

\begin{figure*}
    \centering
    \includegraphics[width=\textwidth]{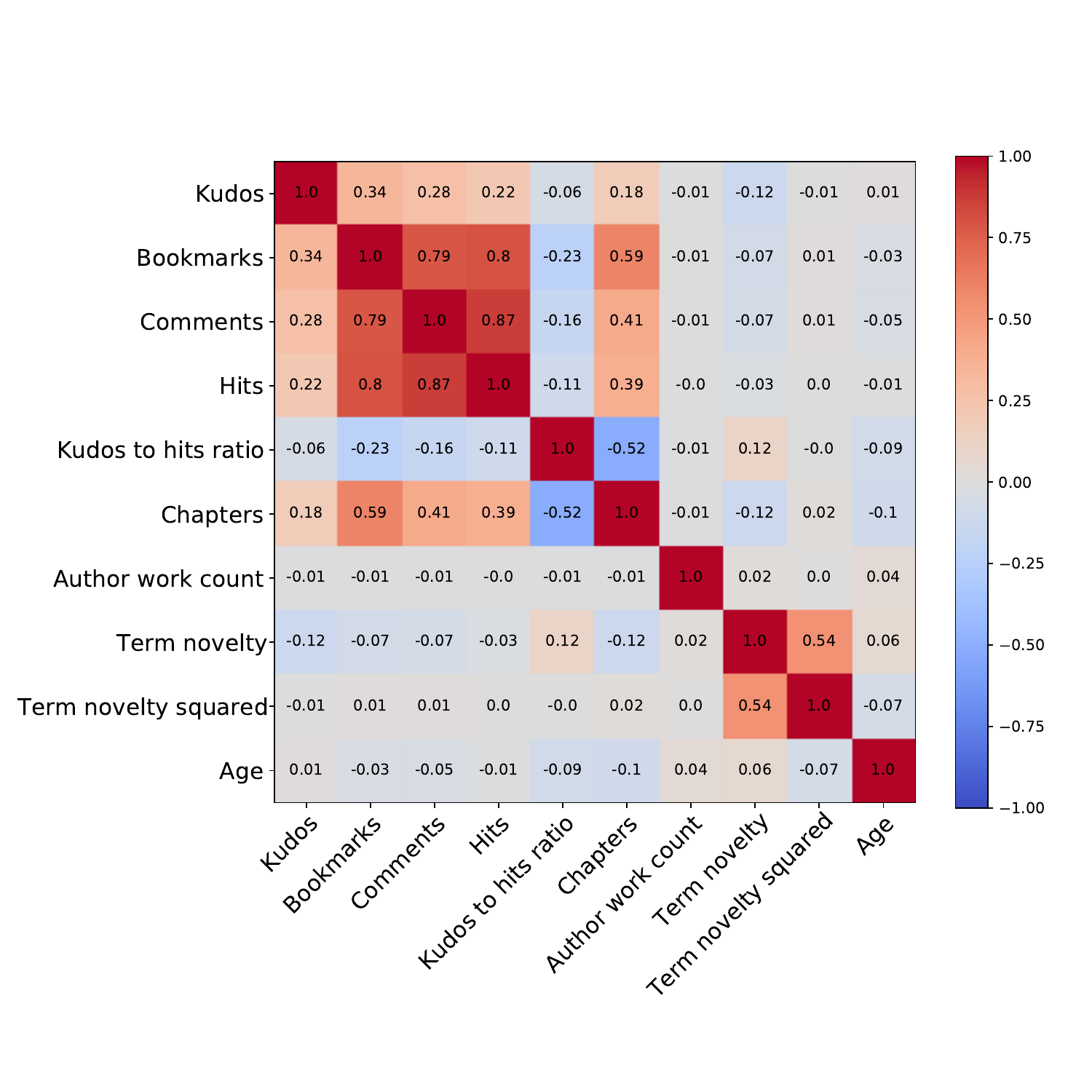}
    \caption{Correlations between the numerical predictor, response, and control variables based on term-level novelty. Strong positive correlation is found between the response variables except for the Kudos-to-hits ratio. Term novelty is weakly negatively correlated with the response variables. Additionally, the number of chapters is moderately positively correlated with success.}
    \label{fig:variables_corr_term_only}
    \end{figure*}

\begin{figure*}
    \centering
    \includegraphics[width=\textwidth]{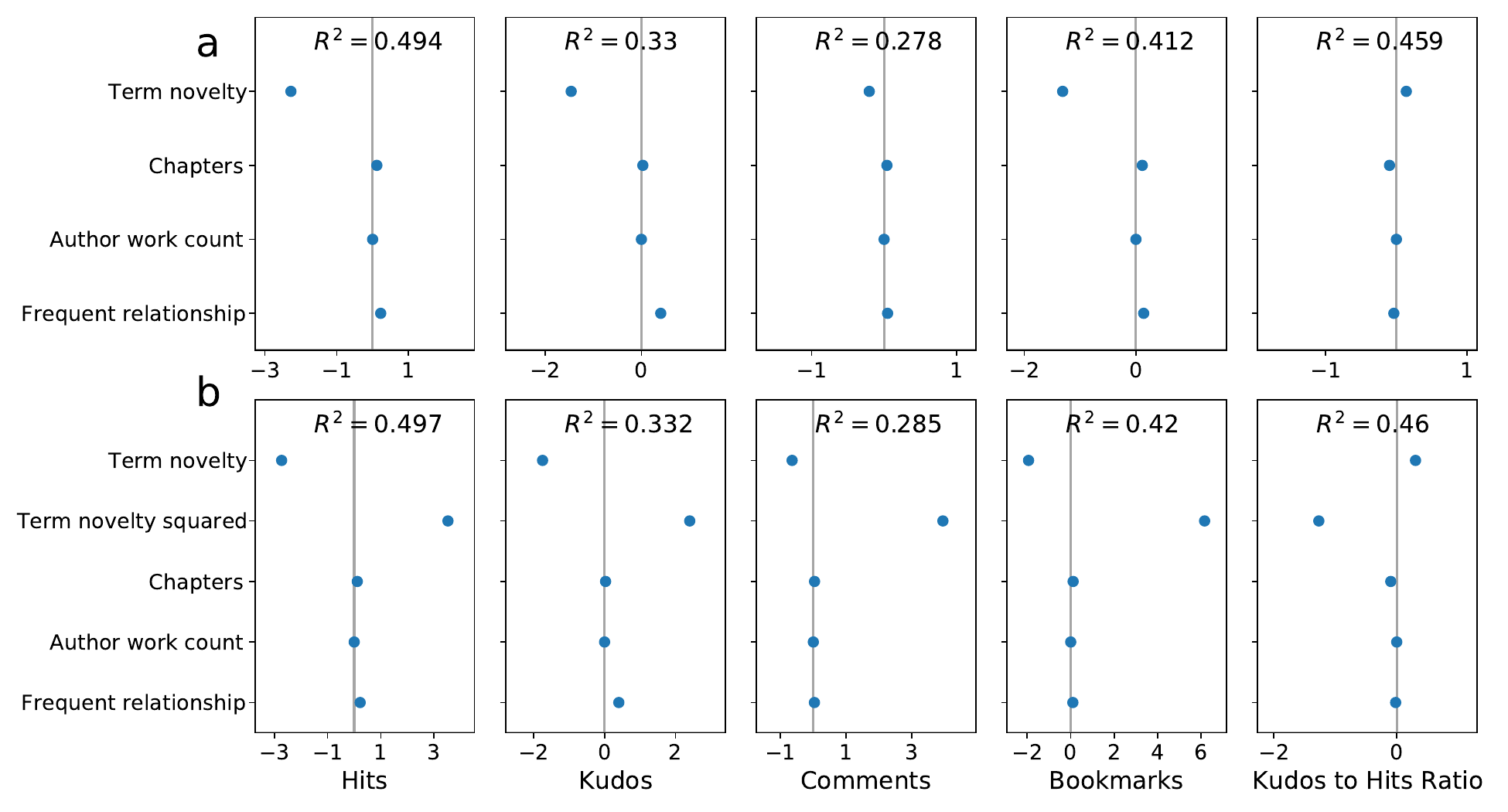}
    \caption{OLS coefficients for the independent variables and selected control variables for the multiple regression models, with term novelty as the focus. $N = 520,730$. Figure (a) does not include the square value of novelty score, while Figure (b) does. 95\% confidence intervals are shown. The coefficients of the categorical variables are omitted. Among the control variables, the number of chapters and the author's fame does not have an effect on the outcome, while featuring a frequent relationship has a positive effect. In both Figure (a) and (b), term novelty has negative effects for Kudos, hits, comments, and bookmarks, but positive for Kudos-to-hits ratio. Figure (b) shows that the squared value of novelty has an opposite effect for each outcome variable.}
    \label{fig:ols_term_only}
    \end{figure*}

\begin{figure*}
    \centering
    \includegraphics[width=\textwidth]{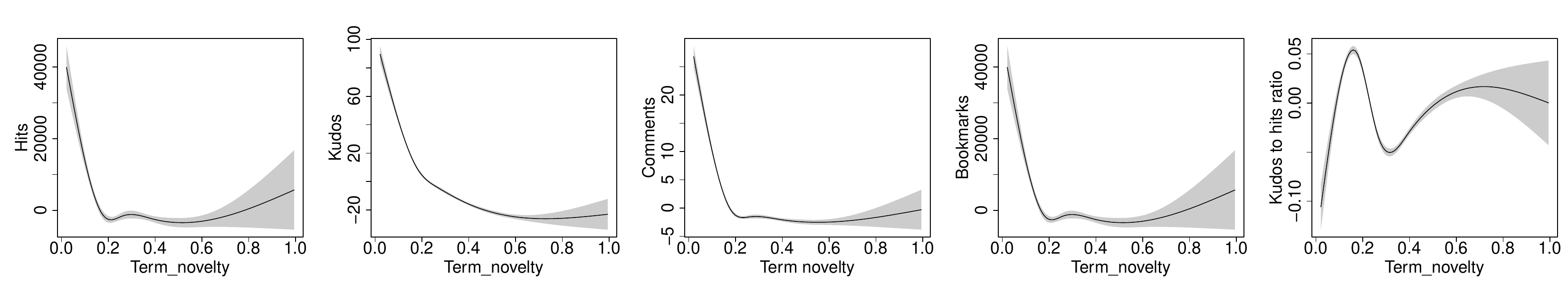}
    \caption{Partial dependence plots of the Generalized Additive Models showing the effect of term novelty. The x-axes show the novelty scores. The y-axes show the partial influence of the novelty score on the response variables while holding other independent variables constant. 95\% confidence intervals are shown.}
    \label{fig:gam_term_only}
    \end{figure*}

    We present results based on this term-level measurement in Figures~\ref{fig:kudos_term_only}--\ref{fig:gam_term_only}. We found a generally similar pattern in the interaction between novelty and success in Figure~\ref{fig:kudos_term_only}; in the multiple regression outcomes, we found that the term novelty also has a negative effect on Kudos, hits, comments, and bookmarks, but has a small positive effect on Kudos-to-hits ratio. The squared values of term novelty has the opposite effect on all outcome variables. In the GAM model, we observed similar U-shaped curves for Kudos, hits, comments, and bookmarks as in the main paper, with decreasing success as novelty increases, but larger uncertainty and small upticks in the high-novelty range. For Kudos-to-hits ratio, we found a bimodal pattern with one peak of Kudos-to-hits ratio at low term novelty, and one at high novelty.

\begin{figure*}
    \centering
    \hspace{1cm}            
    \includegraphics[width=\textwidth]{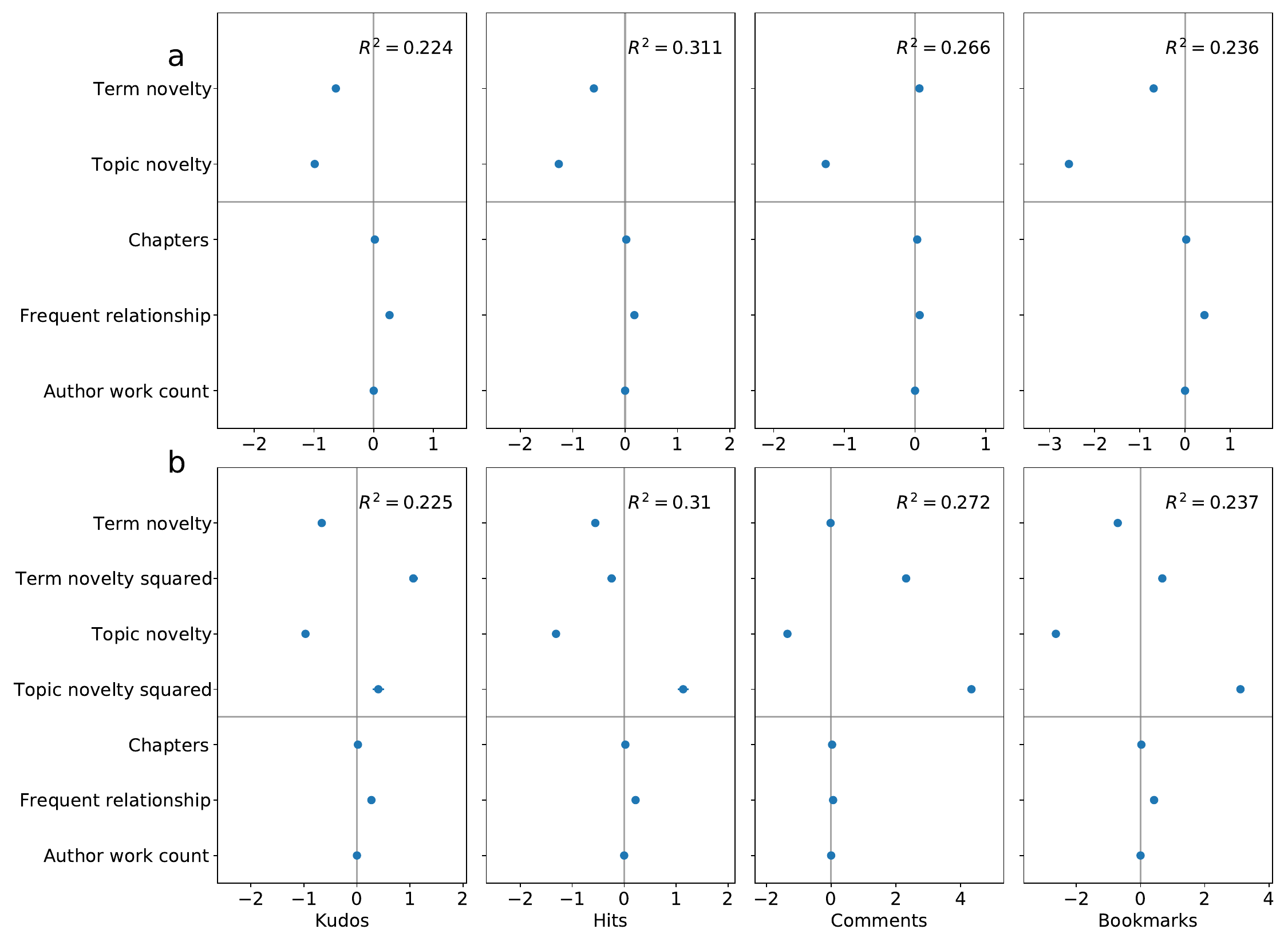}
    \caption{Multiple regression results with pairwise term novelty computation.}
    \label{fig:pairwise_novelty}    
    \end{figure*}

    This approach, similar to the approach in our paper, operationalizes novelty by defining an ``average'' fiction and calculating the distance between each fiction and the average fiction. To remove the potential influence of the average fiction, we experiment computing for each fiction the distance between it and every fiction in the previous time period, and compute the alternative novelty score as:
    
    \begin{equation}
    s_{i}^{(\mathrm{term})} = \frac{1}{N}(\sum_{n=j}^{N}1-\frac{\boldsymbol{f_i} \cdot{\boldsymbol{f_j}}} {\lvert \boldsymbol{f_i} \rvert \lvert \boldsymbol{f_j} \rvert})
    \end{equation}
    
    where $\boldsymbol{f_i}$ is the vector representation of our target fanfiction, and each $\boldsymbol{f_j}$ is a fanfiction from the previous time period. We found that such pairwise definition of novelty does not have a significant influence on our results, as shown in Figure~\ref{fig:pairwise_novelty}. These results support the robustness of our findings against different methods for quantifying novelty.
    
\begin{figure*}
    \centering
    \hspace{1cm}            
    \includegraphics[width=\textwidth]{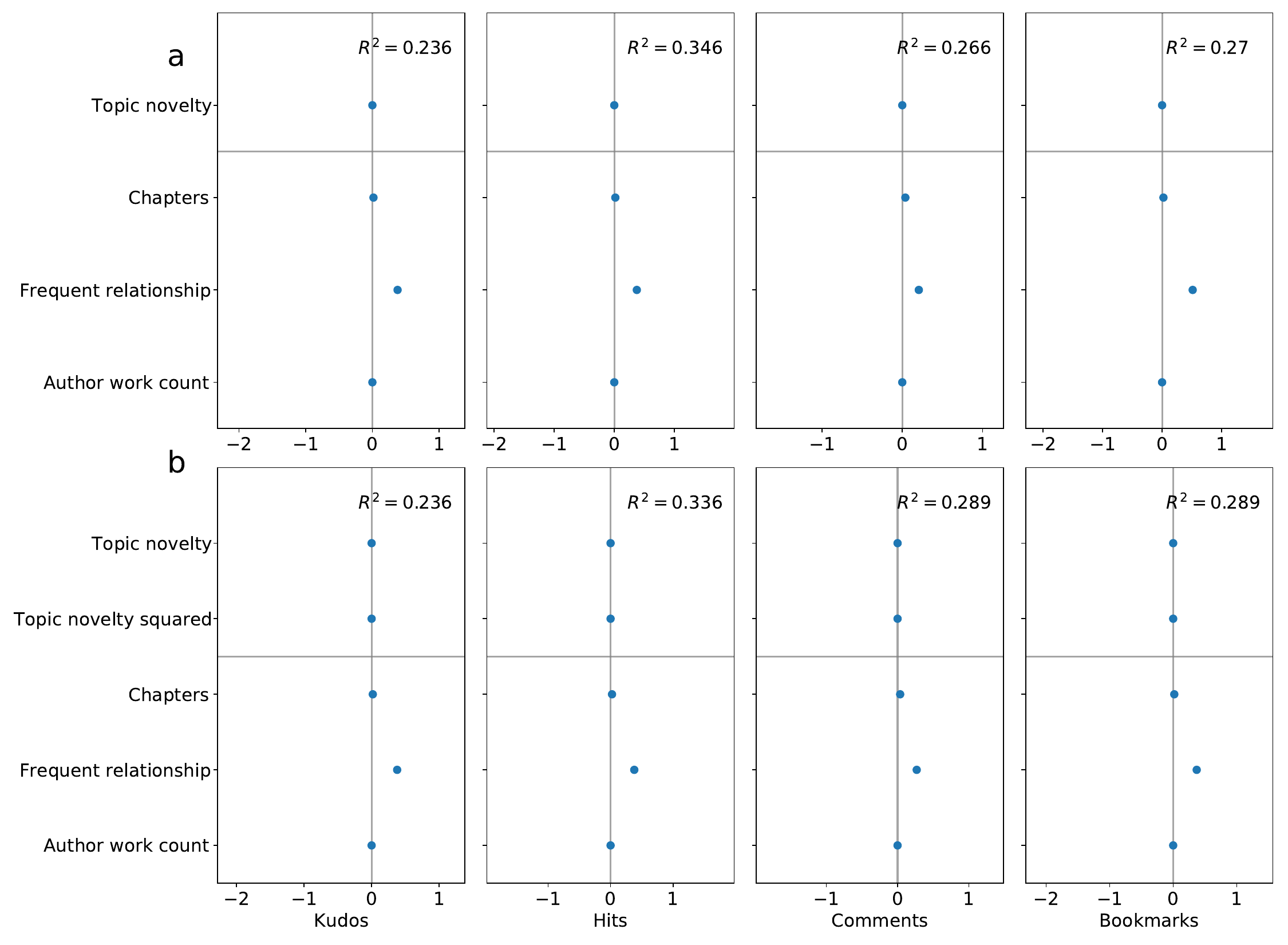}
    \caption{Multiple regression results with perplexity score as topic novelty.}
    \label{fig:perplexity}    
    \end{figure*}

    We also experimented using the topic perplexity score to quantify topic novelty. For fiction published in each month, we train an LDA model on the fiction published within the past 6 months, and use this model to evaluate the perplexity of each of these fiction. A higher perplexity score is interpreted as higher novelty. We found that the perplexity scores have very small effect size in the fitted multiple regression models, as shown in Figure~\ref{fig:perplexity}.

    \subsection*{Full regression results}
\begin{figure*}
    \centering
    \includegraphics[width=0.55\textwidth]{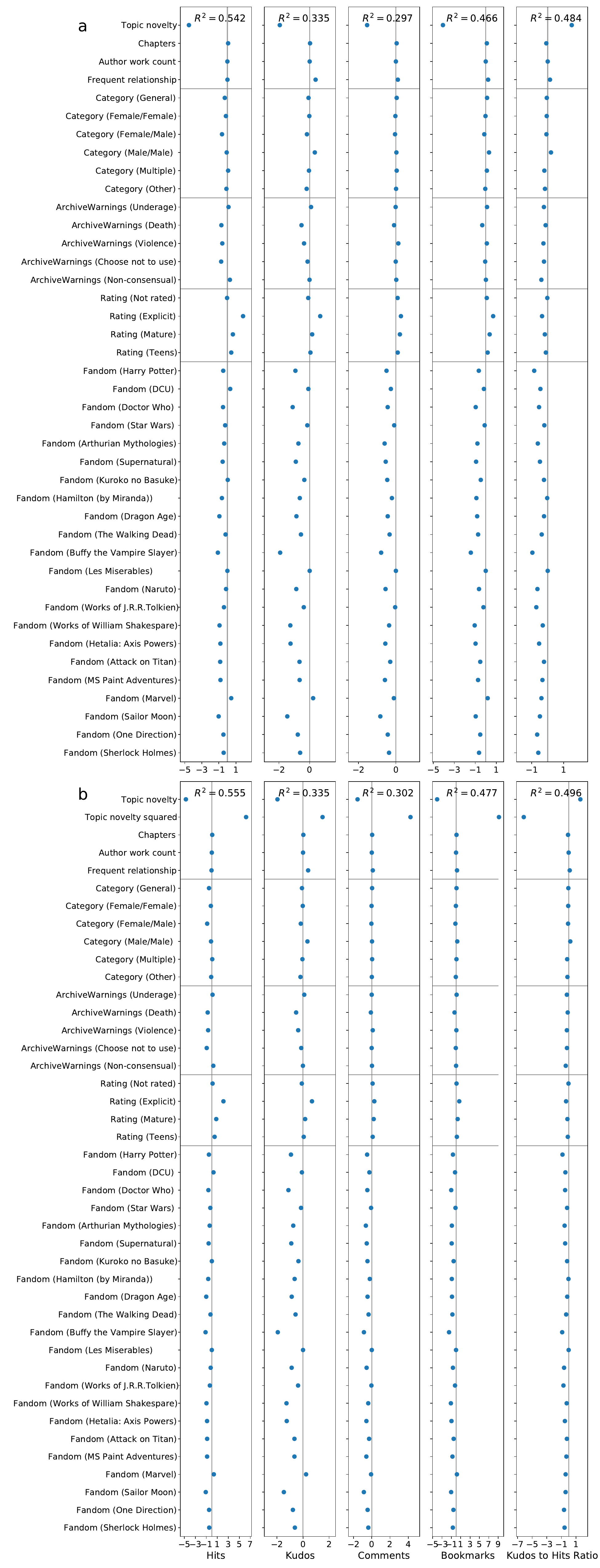}
    \caption{Full multiple regression results with control variables.}
    \label{fig:ols_full}
    \end{figure*}

    Our full multiple regression results with all control variables are shown in Figure~\ref{fig:ols_full}.
    
    \subsection*{KDE results for individual fandoms}
\begin{figure*}
    \centering
    \includegraphics[width=\textwidth]{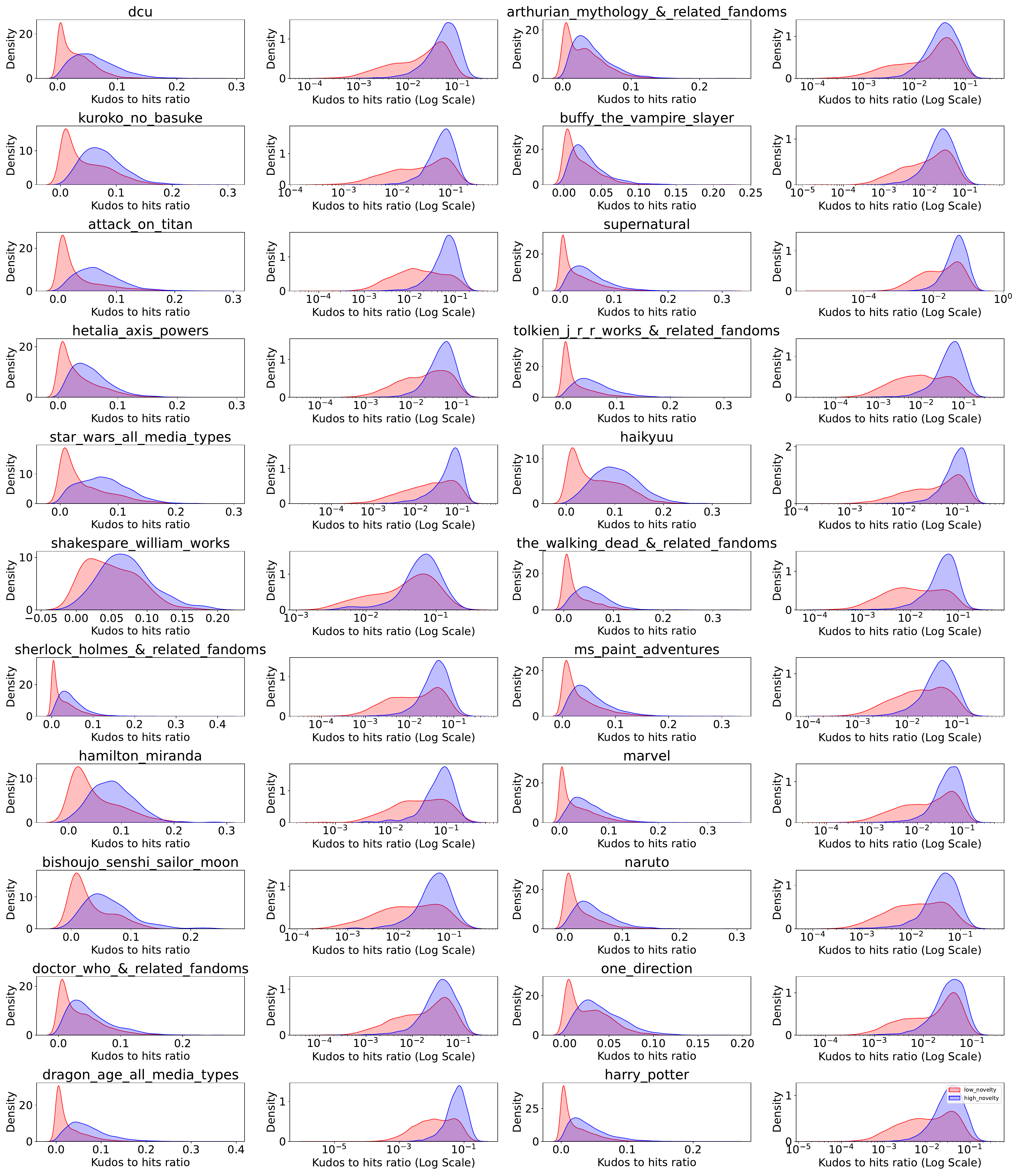}
    \caption{Kernel density estimation (KDE) of the distribution of Kudos to hits ratio for fanfiction in the top 25\% and bottom 25\% novelty range for each fandom plotted separately. Left figure shows KDE in linear scale and right in logarithmic scale for each fandom. High-novelty fanfictions are observed to receive more Kudos with respect to the number of hits they receive, although having fewer Kudos and hits overall.}
    \label{fig:kde_indiv}
    \end{figure*}

    We show the KDE of the distribution of Kudos-to-hits ratio for individual fandoms in Figure~\ref{fig:kde_indiv}. For each fandom, we show the KDE over the top 25\% and botton 25\% of fanfiction ranked by topic novelty. We found the pattern to be consistent over all fandoms.
    
    \FloatBarrier
    
    \FloatBarrier

    \begin{table}
    \centering
    \vspace{-0.5cm}
    \begin{tabular}[width=\textwidth]{p{2cm}p{5cm}}
    \toprule
    Fields & Description \\ 
    \hline
    Title & Title of the work. \\
    Fandoms & The fandom(s) that the work belongs to. \\
    Author & The author(s) of the work.  \\
    Chapters & The number of chapters that the work has. \\
    Archive Warnings & Warnings for sensitive elements. \\
    Category & The type of relationships in the work. \\
    Rating & The age rating. \\
    Relationship & The relationship(s) between characters in the work, in the form of Character A/Character B or Character A\&Character B. \\
    Publish Date & The date the work was published on AO3. \\
    Complete date & For multi-chapter works, the date when it was marked as ``complete''.\\
    Update date & The date when the work was last updated. \\
    \hline
    Kudos & The number of kudos (likes) the work received. \\
    Comments & The number of comments the work received.\\
    Hits & The number of times the link to the work was clicked on. \\
    Bookmarks & The number of people who bookmarked the work.\\
        
    \bottomrule
    \vspace{0.02in}
    \end{tabular}
    \caption{Metadata fields for each work of fanfiction in our dataset.}
    \vspace{-1cm}
    \label{tab:metadata}
    \end{table}%
    \FloatBarrier
    
    \FloatBarrier
    
    \FloatBarrier
    
    \FloatBarrier
    
    \FloatBarrier
    
    \FloatBarrier
    
    \FloatBarrier

    \FloatBarrier

\end{document}